\documentclass{article}
\usepackage{spconf,amsmath,graphicx}
\usepackage{epstopdf}
\usepackage{array}
\usepackage{graphicx}
\usepackage{multirow}
\usepackage{color, soul}
\usepackage{cite}
\usepackage{url}
\usepackage{hyperref}
\usepackage{booktabs}
\usepackage{amsmath}

\usepackage{amssymb}

\title{FCTalker: Fine and Coarse Grained Context Modeling for Expressive Conversational Speech Synthesis}
\name{Yifan Hu$^{ 1}$, Rui Liu$^{ 1,*}$\thanks{*: Corresponding author.}\thanks{This research was funded by the High-level Talents Introduction Project of Inner Mongolia University (No. 10000-22311201/002) and the Young Scientists Fund of the National Natural Science Foundation of China (NSFC) (No. 62206136).}, Guanglai Gao$^{ 1}$, Haizhou Li$^{ 2}$}
\address{ $^1$ Inner Mongolia University, China $^2$ The Chinese University of Hong Kong, Shenzhen, China \\
\small{hyfwalker@163.com, liurui\_imu@163.com, csggl@imu.edu.cn, haizhouli@cuhk.edu.cn}
}
\begin{document}
%
\maketitle
\begin{abstract}
Conversational Text-to-Speech (TTS) aims to synthesis an utterance with the right linguistic and affective prosody in a conversational context. The correlation between the current utterance and the dialogue history at the utterance level was used to improve the expressiveness of synthesized speech. However, the fine-grained information in the dialogue history at the word level also has an important impact on the prosodic expression of an utterance, which has not been well studied in the prior work. Therefore, we propose a novel expressive conversational TTS model, termed as FCTalker, that learn the fine and coarse grained context dependency at the same time during speech generation. Specifically, the FCTalker includes fine and coarse grained encoders to exploit the word and utterance-level context dependency.
To model the word-level dependencies between an utterance and its dialogue history, the fine-grained dialogue encoder is built on top of a dialogue BERT model.
The experimental results show that the proposed method outperforms all baselines and generates more expressive speech that is contextually appropriate.
We release the source code at: \url{https://github.com/walker-hyf/FCTalker}
\end{abstract}
\begin{keywords}
Conversational Text-to-Speech (TTS), Fine and Coarse Grained, Context, Expressive 
\end{keywords}

\vspace{-2mm}
\section{Introduction}
\label{sec:intro}
\vspace{-1mm}

In conversational Text-to-Speech (TTS), we take the speaker interaction history between two speakers into account and generate expressive speech for a target speaker \cite{guo2021conversational,cong2021controllable}. This technique is highly demanded in the deployment of intelligent agents~\cite{mctear2020conversational,Lin_2022}.

With the advent of deep learning,
neural TTS \cite{wang2017tacotron,shen2018natural, ren2019fastspeech,ren2020fastspeech}, i.e. Tacotron \cite{wang2017tacotron,shen2018natural}, FastSpeech \cite{ren2019fastspeech, ren2020fastspeech} based models, has gained remarkable performance over the traditional statistical parametric speech synthesis methods~\cite{zen2009statistical
,ze2013statistical} in terms of speech quality. However, the prosodic rendering of neural TTS in a conversational context remains a challenge.

\begin{figure}[t]
\centering
\centerline{ \includegraphics[width=0.8\linewidth]{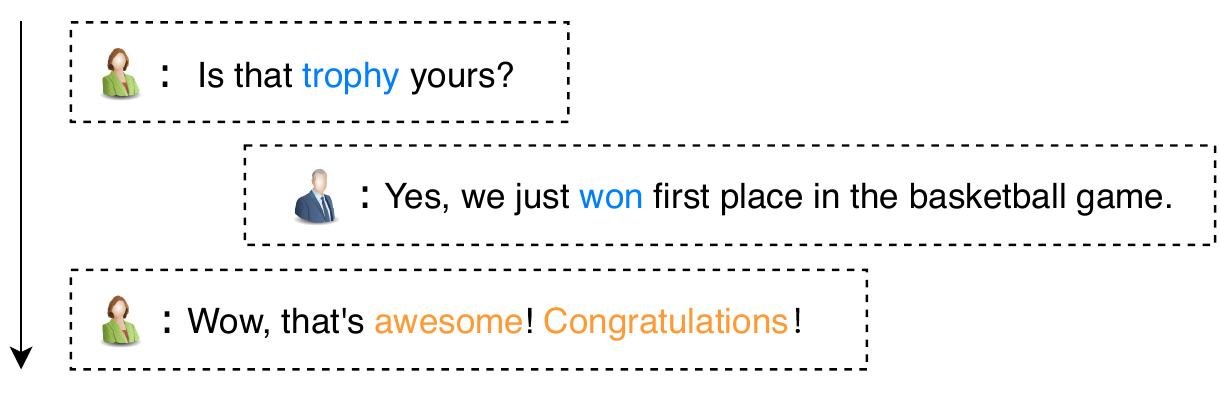}}
\vspace{-3mm}
\caption{An example of word-level context dependencies in a conversation, which the blue words in the conversation history have a direct effect on the prosodic expression of the orange words in the current utterance.}
\label{fig:sample}
\vspace{-5mm}
\end{figure}

The attempts at conversational TTS can be traced back to the HMM era \cite{syrdal2008dialog,koriyama2010conversational,koriyama2011use}.
They make use of rich textual information, such as dialog acts \cite{syrdal2008dialog} and extended context \cite{koriyama2011use} for expressive speech generation. However, these approaches are limited by the need of manual annotation and inadequate dialogue representation of the model.
In the context of neural TTS,
Guo et al. \cite{guo2021conversational} proposed a conversation context encoder based on Tacotron2 model to extract utterance-level prosody-related information from the dialogue history. Cong et al. \cite{cong2021controllable} proposed a context-aware acoustic model which predicting the utterance-level acoustic embedding according to the dialogue history. Mitsui et al. \cite{mitsui2022end} exploited utterance-level BERT encoding to predict conversational speaking styles with spontaneous behavior during TTS synthesis. These studies have advanced the state-of-the-art in conversational TTS. However, they didn't exploit  the word-level information in the dialogue history for the prosody rendering of current utterance. 

Speech prosody is rendered at various segmental level from syllable, lexical word, to sentence~\cite{ruan2022hierarchical,wang2022information}. As shown in Fig\ref{fig:sample}, the blue words ``trophy'' and ``won'' are strong indicators that determine the prosodic expression of the final response. We also find that fine-grained token-level information has played a significant role in conversation-related studies, such as multiturn dialog generation \cite{wang2022information}, conversational emotion recognition \cite{ruan2022hierarchical}, dialogue state tracking \cite{quan2020modeling,kumar2020ma}, conversation intent classification \cite{gangadharaiah2019joint} etc. They simultaneously model the hierarchical contextual semantic dependencies, i.e. word and sentence, between the current utterances and its conversational history, and achieve performance gains.

Inspired by this, we design a novel dialogue BERT \cite{kenton2019bert} based fine-grained encoder in addition to the traditional coarse-grained context encoder. Pre-training a fine-grained encoder, we take the word-level sequences of dialogue history and current utterance as input to learn their fine-grained dependency. At last, the outputs of the fine and coarse-grained context encoders are combined with the phoneme encoding of the input text for expressive speech generation.


The main contributions of this paper include: 1) We propose a novel conversational TTS model FCTalker to synthesis expressive speech; 2) We design a novel fine and coarse grained context modeling scheme to incorporate word and utterance level context; and 3) The proposed model outperforms all state-of-the-art baselines in terms of expressiveness rendering. To our best knowledge, this is the first in-depth conversational TTS study that leverage both word and utterance level information for prosodic rendering. 

\vspace{-3mm}
\section{FCTalker: Methodology }
\label{sec:model}
\vspace{-1mm}
As shown in Fig.\ref{fig:model}, our FCTalker adopts the non-autoregressive Fastspeech2 \cite{ren2020fastspeech} TTS framework as the backbone, which includes fine-grained context encoder, coarse-grained context encoder, text encoder, speaker encoder,  duration predictor, length regulator, variance adaptor and the mel-spectrum decoder.

\vspace{-3mm}
\subsection{Overall Architecture}
\vspace{-1mm}

Given a conversation of $T$ dialogue turns, $L=\{{L_{1}, L_{2}, ...L_{T}}\}$, where $L_{his}=\{{L_{1}, L_{2}, ...L_{T-1}}\}$ is seen as the dialogue history, while  $L_{T}$ is the current utterance to be synthesized. For each utterance $L_{i}$ ($i \in [1,T]$), $W_{i}$ indicates the word-level sequence. Therefore, $W_{his} = \{W_{1}, W_{2}, ..., W_{T-1}\}$ is used to indicate the word-level sequence of dialogue history while $W_{T}$ to represent the word-level sequence of current utterance.

\begin{figure}[t]
\centering
\centerline{ \includegraphics[width=\linewidth]{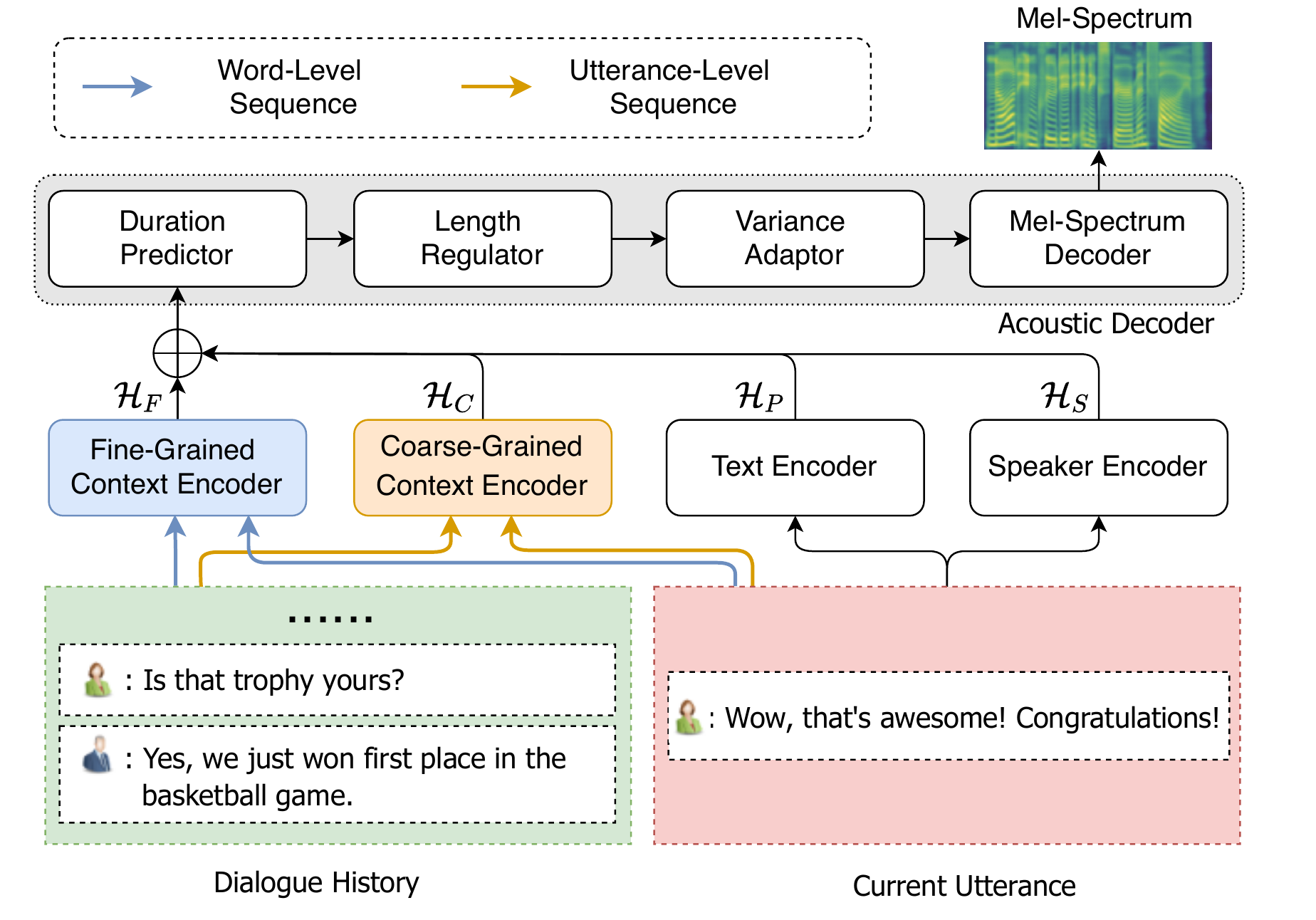}}
\vspace{-3mm}
\caption{The overall architecture of the FCTalker model. The fine-and coarse-grained context encoders aim to learn the word-and utterance-level context dependency respectively between the dialogue history and the current utterance.
}
\vspace{-5mm}
\label{fig:model}
\end{figure}

The fine-grained context encoder $\text{Enc}_{FG}$ takes the $W_{his}$ and $W_{T}$ as input to generate the fine-grained context embedding $\mathcal{H}_{F}$. 
The coarse-grained context encoder $\text{Enc}_{CG}$ reads the dialogue history and current utterance in utterance, that are $L_{his}$ and $L_{T}$, to extract the coarse-grained context embedding $\mathcal{H}_{C}$.
The text encoder, consists of 4 layers of Feed-Forward Transformer (FFT) blocks, seeks to extract the phoneme embedding $\mathcal{H}_{P}$ for the input current utterance. The speaker encoder is a learnable lookup table, that encodes the speaker identity into a speaker {code} $\mathcal{H}_{S}$ \cite{gibiansky2017deep,ping2018deep} to represent the speaker of current utterance.
Similar with the \cite{ren2020fastspeech}, the remaining module, acoustic decoder, consists of duration predictor, length regulator, variance adaptor and the mel-spectrum decoder. 
\textcolor{black}{Variance adaptor aims to estimate different variance information such as duration, pitch and energy and add them into the hidden sequence to predict the mel-spectrum features.}
For duration predictor, we replace the external aligner section of the FastSpeech2 framework with a CTC-based aligner \cite{badlani2022one} since external aligners show a risk of an out-of-distribution problem and perform misalignment for some data.
\textcolor{black}{Note that the fine and coarse grained context embeddings $\mathcal{H}_{F}$, $\mathcal{H}_{C}$ seek to modulate the input phoneme embedding $\mathcal{H}_{P}$ with the word and utterance level context information, which enables all variance information (including pitch, energy and duration) to be affected to express the prosodic in conversation scenario.}
As shown in Fig.\ref{fig:model}, fine-and coarse-grained context embeddings $\mathcal{H}_{F}$, $\mathcal{H}_{C}$, phoneme embedding $\mathcal{H}_{P}$ and the speaker code $\mathcal{H}_{S}$ are concatenated together and feed to the acoustic decoder to predict the mel-spectrum feature. 
We employ HiFi-GAN vocoder \cite{kong2020hifi} to convert the generated mel-spectrum into speech waveform.

We follow \cite{guo2021conversational} to implement the coarse-grained context encoder structure. Each utterance $L_{i} (i\in [1,T])$ is embedded using a pre-trained BERT model to obtain the sentence embedding. Each sentence embedding is appended with a one-hot vector as a speaker ID to distinguish different speakers. 
Then the GRU layer is used to encode the sentence embedding sequence of dialogue history as a prosody feature, which is then concatenated with the sentence embedding of the current utterance. The following linear layer is used to convert the concatenated feature to the final utterance-level context embedding $\mathcal{H}_{C}$ \cite{guo2021conversational}. 
For fine-grained context encoder, to model deep contextual dependencies at the word level, we build its structure with the dialogue BERT model \cite{kenton2019bert}. The parameters of the fine-grained context encoder are initialized by a pre-training step, termed as ``Fine-Grained Context Modeling with Dialogue BERT Pre-training''. We will demonstrate the details of pre-training and fine-grained context embedding extraction in the following subsections.

\vspace{-3mm}
\subsection{Fine-Grained Context Encoder}
 \vspace{-1mm}
\subsubsection{Dialogue BERT Pretrianing}
\label{subsec:pretrain}
 \vspace{-2mm}

We use BERT's deep bidirectional context modeling capability \cite{kenton2019bert} to model the word-level dependencies of dialogue history and current utterance. We use multiple turns of a conversation involving different speakers for the BERT pretraining in this work.

\begin{figure}[t]
\centering
\setlength{\abovecaptionskip}{-0mm}   
\centerline{\includegraphics[width=1\linewidth]{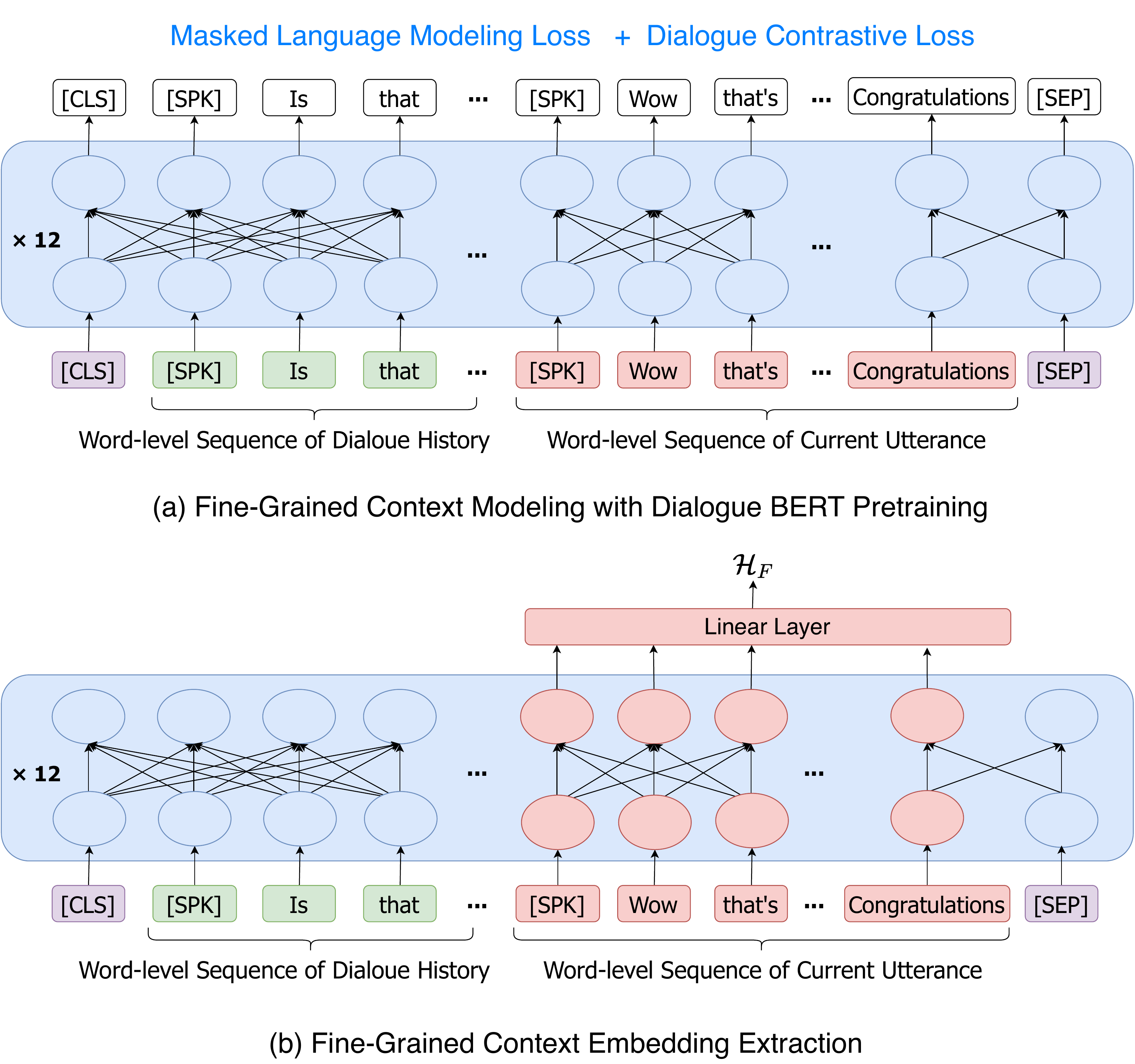}}
\vspace{-4mm}
\caption{The fine-grained context encoder is implemented on the basis of a pretrained dialogue BERT model, as shown in panel (a); we add a additional linear layer on the top of the neurons for current utterance to extract the fine-grained context embedding, as shown in panel (b).}
\label{fig:todbert}
\vspace{-4mm}
\end{figure}

As shown in Fig.\ref{fig:todbert}(a), the input sequence of dialogue BERT based $\text{Enc}_{FG}$ consists of [CLS] token, $W_{his}$ and $W_{T}$ sequence, and the [SEP] token. To capture speaker information and the underlying interaction behavior in conversation, the speaker token [SPK] for each utterance is prefixed to each word-level sequence $W_{i}$. 
At last, the input of the dialogue BERT pre-training model is processed as ``[CLS] [SPK] $W_{1}$ [SPK] $W_{2}$ ... [SPK] $W_{T-1}$  [SPK] $W_{T}$ [SEP]'' with standard positional embeddings and segmentation embeddings.
With the help of global context modeling capacity of self-attention \cite{vaswani2017attention}, the multi-layer Transformer blocks are stacked to learn the deep context for the word-level input sequence. We follow \cite{wu2020tod} and build the dialogue BERT model with BERT-based uncased model \cite{wu2020tod}, which includes 12-layers Transformer blocks with 12 attention heads with hidden sizes $s_{h}=768$ in each layer.

We conduct the pre-training using two loss functions: 1) \textit{Masked Language Modeling Loss ($\mathcal{L}_{mlm}$)} and 2) \textit{Dialogue Contrastive Loss ($\mathcal{L}_{dc}$)}. $\mathcal{L}_{mlm}$ is a common objective function for BERT-like architectures. Note that unlike the BERT model that mask and replace the token once before training \cite{kenton2019bert}, inspired by \cite{wu2020tod}, we conduct token masking dynamically during batch training. The $\mathcal{L}_{mlm}$ is defined as: $ \mathcal{L}_{m l m}=-\sum_{m=1}^M \log P\left(x_m\right)$, where $M$ is the total number of masked tokens and $P(x_{m})$ is the predicted probability of the token $x_{m}$ over the vocabulary size. $\mathcal{L}_{dc}$ encourage the model to capture the underlying conversational sequential order, context information, etc. Suppose that there are $T'$ conversations, each consisting of multiple turns of dialogue. The $\mathcal{L}_{dc}$ is defined as: $\mathcal{L}_{d c}=-\sum_{i=1}^{T'} \log M_{k, k} ; \quad \!\!\!\!\!\! M=\operatorname{Softmax}(C \!R^\top ) \in \mathbb{R}^{T' \times T'}$, where $C\in \mathbb{R}^{T' \times s_{h}}$ and $R\in \mathbb{R}^{T' \times s_{h}}$ are history and current matrices respectively by taking the output [CLS] representations from the $T'$ dialogues. More details are refer to \cite{wu2020tod}.
At last, we sum them up as the total loss: $\mathcal{L}$ = $\mathcal{L}_{mlm}$ + $\mathcal{L}_{dc}$.

\vspace{-3mm}

\vspace{-2mm}
\subsubsection{Fine-Grained Context Embedding Extraction}
\vspace{-2mm}
During the FCTalker training, the $\text{Enc}_{FG}$ learns the word-level context information to extract the context embedding for current utterance. 
As shown in Fig.\ref{fig:todbert}(b), we add a linear layer on the top of the neurons for $W_{T}$ to extract the fine-grained context embedding $\mathcal{H}_{F}$. The neurons for dialogue history and other special tokens are discarded.

There was a study \cite{li_inferring_2022} of a local attention module to learn the fine-grained context in a conversation. For clarity, we would like to highlight two differences of our work: 1) \cite{li_inferring_2022} is focused on the multimodal conversation while this work is only on  textual modality;
2) the simple attention mechanism in \cite{li_inferring_2022} only learns the interaction and ignores the deep semantic dependency between the dialogue history and the current utterance, that we learn via a BERT model in this study. 


\vspace{-5mm}
\section{Experiments and Results}
\label{sec:exp}
\vspace{-3mm}
\subsection{Dataset}
\vspace{-2mm}
We validate the FCTalker on a recently public dataset for conversational TTS called DailyTalk \cite{lee2022dailytalk}, which is a subset of the open-domain conversation dataset DailyDialog \cite{li2017dailydialog}. 
DailyTalk consists of 23,773 audio clips representing 20 hours in total, in which 2,541 conversations were sampled, modified, and recorded. All dialogues are long enough to represent the context of each conversation. The dataset was recorded by a male and a female simultaneously. All speech samples are sampled at 44.10 kHz and coded in 16 bits. We partition the speech data into training, validation, and test sets at a ratio of 8:1:1.

The dialogue BERT pretraining is conducted on a set of dialogue corpora \cite{wu2020tod} consisting of nine different task-oriented datasets, such as MetaLWOZ \cite{lee2019multi-domain} and Schema \cite{https://doi.org/10.48550/arxiv.1909.05855}. There are about 100 thousand dialogues covering more than 60 different domains.
\vspace{-4mm}
\subsection{Experimental Setup}
\vspace{-2mm}
The fine-grained encoder consists of 12-layers Transformer blocks and 12 attention heads with hidden size $s_h$ = 768. Note that the additional linear layer of Fig.\ref{fig:todbert}(b) reduces the output dimension from 768 to 256.
The coarse-grained encoder is implemented by following \cite{guo2021conversational}.
The text encoder takes 256-dimensions phoneme sequence as input. We extract 80-channel mel-spectrum features with a frame size of 50ms and 12.5ms frame shift as the reference target. 
{The dimensions of $\mathcal{H}_{F}$, $\mathcal{H}_{C}$, $\mathcal{H}_{P}$ and $\mathcal{H}_{S}$ are all 256}.
The detailed setup of acoustic decoder can be found in \cite{lee2022dailytalk}.

The number of conversation turns $T$ is a hyper-parameter that can be adjusted to validate the effects of different context Windows. We set the value of $T$ to range from 1 to 14 for comparison.
For FCTalker training, we use Adam optimizer with $\beta_{\mbox{\scriptsize 1}}$ = 0.9, $\beta_{\mbox{\scriptsize 2}}$ = 0.98. All speech samples are re-sampled to 22.05 kHZ. The model is trained on a Tesla V100 GPU with a batch size of 32 and {900k} steps. We follow \cite{wu2020tod} to pretrain the fine-grained dialogue BERT with {900k} steps. Due to the space limit, we report the subjective evaluations, instead of objective evaluations for pitch, energy and duration, in following subsection.


\vspace{-4mm}
\subsection{Comparative Study}
\vspace{-2mm}
We develop three neural TTS systems for a comparative study, that include the 1) \textbf{FastSpeech2} \cite{ren2020fastspeech}: the state-of-the-art neural TTS model that takes the single utterance as input, without any conversational context modeling; 2) \textbf{DailyTalk} \cite{lee2022dailytalk}: a latest neural conversational TTS model with coarse-grained context encoder; 3) \textbf{FCTalker}: the proposed model with fine-and coarse-grained context modeling strategy and 4) \textbf{Ground Truth} speech under conversation scenario.

\begin{table}[t]
\centering
\caption{\label{tab:mos_1}The MOS results for all comparative systems,  with 95\% confidence interval.}
\label{tab:mos-1}
\begin{tabular}{lcl}
\toprule
\multirow{2}{*}{\qquad System} & \multicolumn{2}{c}{MOS}     \\ \cline{2-3}& Utterance-level & Dialogue-level \\ \hline
FastSpeech2 \cite{ren2020fastspeech} & 3.80 $\pm$ 0.026  & 3.90 $\pm$ 0.046 \\  
DailyTalk \cite{lee2022dailytalk} ($T$=2)& 3.86 $\pm$ 0.042  & 3.96 $\pm$ 0.056 \\ \hline
\textbf{FCTalker} ($T$=2)  & \textbf{3.97 $\pm$ 0.040}   & \textbf{4.15 $\pm$ 0.031}   \\ \hline
\textbf{Ground Truth}   & \textbf{4.46 $\pm$ 0.031}     & \textbf{4.52 $\pm$ 0.034} \\ \bottomrule
\end{tabular}
\vspace{-5mm}
\end{table}

\vspace{-4mm}
\subsection{Experiments and Results}
\vspace{-2mm}
We conduct 5-point Likert scale mean opinion score (MOS) \cite{streijl2016mean} listening experiments \footnote{Speech samples: \url{https://walker-hyf.github.io/FCTalker/}} to validate the FCTalker model in terms of prosodic expressiveness and context modeling.

\vspace{-4mm}
\subsubsection{Prosodic Expressiveness Evaluation}
\label{sec:mos1}
\vspace{-2mm}
Note that the prosodic expressiveness of conversational speech is related to its adjacent dialogue, therefore, we design two kinds of experiments, including \textbf{Utterance-level} and \textbf{Dialogue-level MOS} tests, to test the prosodic performance of synthesized speech in a conversation scenario. Specifically, \textit{Utterance-level MOS} means the volunteers only rated single sentences from different systems, without reference to the dialogue history. \textit{Dialogue-level MOS} indicates that volunteers rate a multi-turn conversation that includes several audios, some of which are synthesized from the TTS model.


In this section, each audio is listened by 30 volunteers, each of which listens to 80 speech samples. For DailyTalk and FCTalker, we set the dialogue turns $T$ to 2 and report the results in Table \ref{tab:mos-1}.
It's observed that our FCTalker outperforms all baselines and performs the smallest gap with the ground truth speech, in both the utterance and dialogue scenarios. FastSpeech2 achieves the lowest performance due to the lack of prosodic information in context during speech generation. 
DailyTalk can produce more prosodic speech than FastSpeech2 since it incorporates utterance-level contextual information.
As expected, FCTalker achieves the best performance by considering both word-level and sentence-level contextual information, which is remarkable.


\begin{figure}[t]
\centering
\setlength{\abovecaptionskip}{-0mm}   
\centerline{\includegraphics[width=1\linewidth]{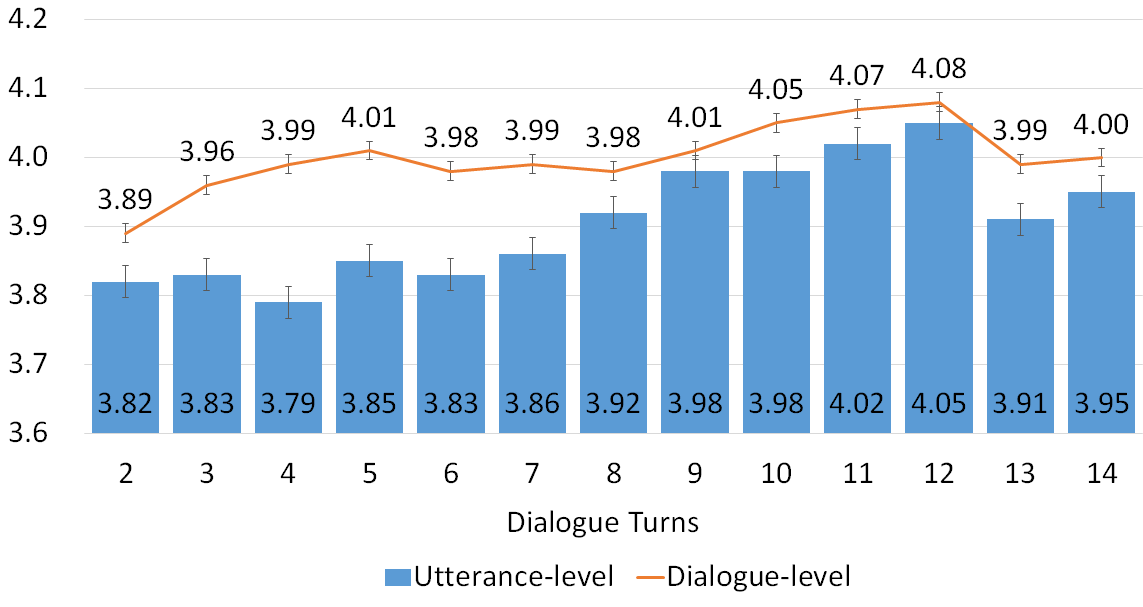}}
\vspace{-5mm}
\caption{The MOS results of various dialogue history turn number $T$ (range from 2 to 14) for FCTalker, with 95\% confidence interval.}
\label{fig:turns}
\vspace{-5mm}
\end{figure}

\vspace{-4mm}
\subsubsection{Dialogue History Turns Comparison}
\vspace{-2mm}
Contextual information is important for the prosodic expression of the current statement, but how the length of a dialog history turn is chosen to achieve the optimal effect needs to be further verified. In this section, considering the  average number 9.3 of dialogue turn in the DailyTalk, we set the dialogue turn $T$ range from 2 to 14 to compare the utterance-and dialogue-level MOS scores.

We also invite 30 volunteers and follow the settings of Sec.\ref{sec:mos1} to conduct the listening test.
As shown in Fig.\ref{fig:turns}, when the value of dialogue turn $T$ increases from 2 to 12, the utterance-and dialogue-level MOS scores show an overall upward trend, while they show a certain downward trend when it increases from 12 to 14.
We also found that when the number of $T$ ranged from 9 to 12, the MOS scores remained above 4.0 steadily. Given that the average dialogue turn of DailyTalk is 9.3, we concluded that it might be a good choice to set the optimal $T$ according to the average dialogue turn.

\vspace{-4mm}
\section{Conclusion}
\label{sec:con}
\vspace{-3mm}
This work proposed a novel expressive conversational TTS model, termed as FCTalker, which includes both fine-and coarse-grained encoders. 
The fine-grained encoder is implemented on a pre-trained dialogue BERT model and can learn the deep fine-grained context dependencies of the dialogue history and the current utterance. Experimental results show that our model outperforms all baselines and generates expressive speech that is more in line with the conversational context. We also give suggestions on the selection of the optimal dialogue history turns through further analysis. To further optimize the context modeling mechanism  will be our future focus.





\bibliographystyle{IEEEbib}
{\footnotesize
\bibliography{strings}}

\begin{thebibliography}{10}

\bibitem{guo2021conversational}
Haohan Guo, Shaofei Zhang, Frank~K Soong, Lei He, and Lei Xie,
\newblock ``Conversational end-to-end tts for voice agents,''
\newblock in {\em 2021 IEEE Spoken Language Technology Workshop (SLT)}. IEEE,
  2021, pp. 403--409.

\bibitem{cong2021controllable}
Jian Cong, Shan Yang, Na~Hu, Guangzhi Li, Lei Xie, and Dan Su,
\newblock ``Controllable context-aware conversational speech synthesis,''
\newblock {\em arXiv preprint arXiv:2106.10828}, 2021.

\bibitem{mctear2020conversational}
Michael McTear,
\newblock ``Conversational ai: Dialogue systems, conversational agents, and
  chatbots,''
\newblock {\em Synthesis Lectures on Human Language Technologies}, vol. 13, no.
  3, pp. 1--251, 2020.

\bibitem{Lin_2022}
Ting-En Lin, Yuchuan Wu, Fei Huang, Luo Si, Jian Sun, and Yongbin Li,
\newblock ``Duplex conversation: Towards human-like interaction in spoken
  dialogue systems,''
\newblock in {\em Proceedings of the 28th {ACM} {SIGKDD} Conference on
  Knowledge Discovery and Data Mining}. aug 2022, {ACM}.

\bibitem{wang2017tacotron}
Yuxuan Wang, RJ~Skerry-Ryan, Daisy Stanton, Yonghui Wu, Ron~J Weiss, Navdeep
  Jaitly, Zongheng Yang, Ying Xiao, Zhifeng Chen, Samy Bengio, et~al.,
\newblock ``Tacotron: Towards end-to-end speech synthesis,''
\newblock {\em arXiv preprint arXiv:1703.10135}, 2017.

\bibitem{shen2018natural}
Jonathan Shen, Ruoming Pang, Ron~J Weiss, Mike Schuster, Navdeep Jaitly,
  Zongheng Yang, Zhifeng Chen, Yu~Zhang, Yuxuan Wang, Rj~Skerrv-Ryan, et~al.,
\newblock ``Natural tts synthesis by conditioning wavenet on mel spectrogram
  predictions,''
\newblock in {\em 2018 IEEE international conference on acoustics, speech and
  signal processing (ICASSP)}. IEEE, 2018, pp. 4779--4783.

\bibitem{ren2019fastspeech}
Yi~Ren, Yangjun Ruan, Xu~Tan, Tao Qin, Sheng Zhao, Zhou Zhao, and Tie-Yan Liu,
\newblock ``Fastspeech: Fast, robust and controllable text to speech,''
\newblock {\em Advances in Neural Information Processing Systems}, vol. 32,
  2019.

\bibitem{ren2020fastspeech}
Yi~Ren, Chenxu Hu, Xu~Tan, Tao Qin, Sheng Zhao, Zhou Zhao, and Tie-Yan Liu,
\newblock ``Fastspeech 2: Fast and high-quality end-to-end text to speech,''
\newblock {\em arXiv preprint arXiv:2006.04558}, 2020.

\bibitem{zen2009statistical}
Heiga Zen, Keiichi Tokuda, and Alan~W Black,
\newblock ``Statistical parametric speech synthesis,''
\newblock {\em speech communication}, vol. 51, no. 11, pp. 1039--1064, 2009.

\bibitem{ze2013statistical}
Heiga Ze, Andrew Senior, and Mike Schuster,
\newblock ``Statistical parametric speech synthesis using deep neural
  networks,''
\newblock in {\em 2013 ieee international conference on acoustics, speech and
  signal processing}. IEEE, 2013, pp. 7962--7966.

\bibitem{syrdal2008dialog}
Ann~K Syrdal and Yeon-Jun Kim,
\newblock ``Dialog speech acts and prosody: Considerations for tts,''
\newblock in {\em Proceedings of speech prosody}, 2008, pp. 661--665.

\bibitem{koriyama2010conversational}
Tomoki Koriyama, Takashi Nose, and Takao Kobayashi,
\newblock ``Conversational spontaneous speech synthesis using average voice
  model,''
\newblock in {\em Eleventh Annual Conference of the International Speech
  Communication Association}, 2010.

\bibitem{koriyama2011use}
Tomoki Koriyama, Takashi Nose, and Takao Kobayashi,
\newblock ``On the use of extended context for hmm-based spontaneous
  conversational speech synthesis,''
\newblock in {\em Twelfth Annual Conference of the International Speech
  Communication Association}, 2011.

\bibitem{mitsui2022end}
Kentaro Mitsui, Tianyu Zhao, Kei Sawada, Yukiya Hono, Yoshihiko Nankaku, and
  Keiichi Tokuda,
\newblock ``End-to-end text-to-speech based on latent representation of
  speaking styles using spontaneous dialogue,''
\newblock {\em arXiv preprint arXiv:2206.12040}, 2022.

\bibitem{ruan2022hierarchical}
Yu-Ping Ruan, Shu-Kai Zheng, Taihao Li, Fen Wang, and Guanxiong Pei,
\newblock ``Hierarchical and multi-view dependency modelling network for
  conversational emotion recognition,''
\newblock in {\em ICASSP 2022-2022 IEEE International Conference on Acoustics,
  Speech and Signal Processing (ICASSP)}. IEEE, 2022, pp. 7032--7036.

\bibitem{wang2022information}
Jiamin Wang, Xiao Sun, Qian Chen, and Meng Wang,
\newblock ``Information-enhanced hierarchical self-attention network for
  multiturn dialog generation,''
\newblock {\em IEEE Transactions on Computational Social Systems}, 2022.

\bibitem{quan2020modeling}
Jun Quan and Deyi Xiong,
\newblock ``Modeling long context for task-oriented dialogue state
  generation,''
\newblock {\em arXiv preprint arXiv:2004.14080}, 2020.

\bibitem{kumar2020ma}
Adarsh Kumar, Peter Ku, Anuj Goyal, Angeliki Metallinou, and Dilek Hakkani-Tur,
\newblock ``Ma-dst: Multi-attention-based scalable dialog state tracking,''
\newblock in {\em Proceedings of the AAAI Conference on Artificial
  Intelligence}, 2020, vol.~34, pp. 8107--8114.

\bibitem{gangadharaiah2019joint}
Rashmi Gangadharaiah and Balakrishnan Narayanaswamy,
\newblock ``Joint multiple intent detection and slot labeling for goal-oriented
  dialog,''
\newblock in {\em Proceedings of the 2019 Conference of the North American
  Chapter of the Association for Computational Linguistics: Human Language
  Technologies, Volume 1 (Long and Short Papers)}, 2019, pp. 564--569.

\bibitem{kenton2019bert}
Jacob Devlin, Ming-Wei Chang, Kenton Lee, and Kristina Toutanova,
\newblock ``Bert: Pre-training of deep bidirectional transformers for language
  understanding,''
\newblock in {\em Proceedings of NAACL-HLT}, 2019, pp. 4171--4186.

\bibitem{gibiansky2017deep}
Andrew Gibiansky, Sercan Arik, Gregory Diamos, John Miller, Kainan Peng, Wei
  Ping, Jonathan Raiman, and Yanqi Zhou,
\newblock ``Deep voice 2: Multi-speaker neural text-to-speech,''
\newblock {\em Advances in neural information processing systems}, vol. 30,
  2017.

\bibitem{ping2018deep}
Wei Ping, Kainan Peng, Andrew Gibiansky, Sercan~O Arik, Ajay Kannan, Sharan
  Narang, Jonathan Raiman, and John Miller,
\newblock ``Deep voice 3: Scaling text-to-speech with convolutional sequence
  learning,''
\newblock in {\em International Conference on Learning Representations}, 2018.

\bibitem{badlani2022one}
Rohan Badlani, Adrian {\L}a{\'n}cucki, Kevin~J Shih, Rafael Valle, Wei Ping,
  and Bryan Catanzaro,
\newblock ``One tts alignment to rule them all,''
\newblock in {\em ICASSP 2022-2022 IEEE International Conference on Acoustics,
  Speech and Signal Processing (ICASSP)}. IEEE, 2022, pp. 6092--6096.

\bibitem{kong2020hifi}
Jungil Kong, Jaehyeon Kim, and Jaekyoung Bae,
\newblock ``Hifi-gan: Generative adversarial networks for efficient and high
  fidelity speech synthesis,''
\newblock {\em Advances in Neural Information Processing Systems}, vol. 33, pp.
  17022--17033, 2020.

\bibitem{vaswani2017attention}
Ashish Vaswani, Noam Shazeer, Niki Parmar, Jakob Uszkoreit, Llion Jones,
  Aidan~N Gomez, {\L}ukasz Kaiser, and Illia Polosukhin,
\newblock ``Attention is all you need,''
\newblock {\em Advances in neural information processing systems}, vol. 30,
  2017.

\bibitem{wu2020tod}
Chien-Sheng Wu, Steven~C.H. Hoi, Richard Socher, and Caiming Xiong,
\newblock ``{TOD}-{BERT}: Pre-trained natural language understanding for
  task-oriented dialogue,''
\newblock in {\em Proceedings of the 2020 Conference on Empirical Methods in
  Natural Language Processing (EMNLP)}, Online, Nov. 2020, pp. 917--929.

\bibitem{li_inferring_2022}
Jingbei Li, Yi~Meng, Xixin Wu, Zhiyong Wu, Jia Jia, Helen Meng, Tian Qiao,
  Yuping Wang, and Yuxuan Wang,
\newblock ``Inferring speaking styles from multi-modal conversational context
  by multi-scale relational graph convolutional networks,''
\newblock in {\em Proceedings of the 30th {ACM} {International} {Conference} on
  {Multimedia}}, Lisboa, Portugal, 2022, {MM} '22, ACM.

\bibitem{lee2022dailytalk}
Keon Lee, Kyumin Park, and Daeyoung Kim,
\newblock ``Dailytalk: Spoken dialogue dataset for conversational
  text-to-speech,''
\newblock {\em arXiv preprint arXiv:2207.01063}, 2022.

\bibitem{li2017dailydialog}
Yanran Li, Hui Su, Xiaoyu Shen, Wenjie Li, Ziqiang Cao, and Shuzi Niu,
\newblock ``Dailydialog: A manually labelled multi-turn dialogue dataset,''
\newblock {\em arXiv preprint arXiv:1710.03957}, 2017.

\bibitem{lee2019multi-domain}
Sungjin Lee, Hannes Schulz, Adam Atkinson, Jianfeng Gao, Kaheer Suleman, Layla
  El~Asri, Mahmoud Adada, Minlie Huang, Shikhar Sharma, Wendy Tay, and Xiujun
  Li,
\newblock ``Multi-domain task-completion dialog challenge,''
\newblock in {\em Dialog System Technology Challenges 8}, March 2019.

\bibitem{https://doi.org/10.48550/arxiv.1909.05855}
Abhinav Rastogi, Xiaoxue Zang, Srinivas Sunkara, Raghav Gupta, and Pranav
  Khaitan,
\newblock ``Towards scalable multi-domain conversational agents: The
  schema-guided dialogue dataset,'' 2019.

\bibitem{streijl2016mean}
Robert~C Streijl, Stefan Winkler, and David~S Hands,
\newblock ``Mean opinion score (mos) revisited: methods and applications,
  limitations and alternatives,''
\newblock {\em Multimedia Systems}, vol. 22, no. 2, pp. 213--227, 2016.

\end{thebibliography}

\end{document}